# Exclusion of GNSS NLOS Receptions Caused by Dynamic Objects in Heavy Traffic Urban Scenarios Using Real-Time 3D Point Cloud: An Approach without 3D Maps

Weisong Wen
Department of Mechanical Engineering
The Hong Kong Polytechnic University
Kowloon, Hong Kong
17902061r@connect.polyu.hk

Guohao Zhang, Li-Ta Hsu*
Interdisciplinary Division of Aeronautical and Aviation Engineering
The Hong Kong Polytechnic University
Kowloon, Hong Kong
lt.hsu@polyu.edu.hk

*Abstract*—Absolute positioning is an essential factor for the arrival of autonomous driving. Global Navigation Satellites System (GNSS) receiver provides absolute localization for it. GNSS solution can provide satisfactory positioning in open or sub-urban areas, however, its performance suffered in super-urbanized area due to the phenomenon which are well-known as multipath effects and NLOS receptions. The effects dominate GNSS positioning performance in the area. The recent proposed 3D map aided (3DMA) GNSS can mitigate most of the multipath effects and NLOS receptions caused by buildings based on 3D city models. However, the same phenomenon caused by moving objects in urban area is currently not modelled in the 3D geographic information system (GIS). Moving objects with tall height, such as the double-decker bus, can also cause NLOS receptions because of the blockage of GNSS signals by surface of objects. Therefore, we present a novel method to exclude the NLOS receptions caused by double-decker bus in highly urbanized area, Hong Kong. To estimate the geometry dimension and orientation relative to GPS receiver, a Euclidean cluster algorithm and a classification method are used to detect the double-decker buses and calculate their relative locations. To increase the accuracy and reliability of the proposed NLOS exclusion method, an NLOS exclusion criterion is proposed to exclude the blocked satellites considering the elevation, signal noise ratio (SNR) and horizontal dilution of precision (HDOP). Finally, GNSS positioning is estimated by weighted least square (WLS) method using the remaining satellites after the NLOS exclusion. A static experiment was performed near a double-decker bus stop in Hong Kong, which verified the effectiveness of the proposed method.

## I. INTRODUCTION

Autonomous vehicle [1] is believed to be a remedy to reduce the excessive traffic jams and accidents. To achieve fully autonomous driving in highly urbanized area, absolute lane-level positioning is required. Light detection and ranging (LiDAR), camera and inertial navigation system (INS) are usually integrated with GNSS positioning [2-4]. However, the three positioning sources can only conduct relative positioning. GNSS solution is the only one that can constantly provide absolute positioning and possesses increased popularity because of the availability of multi-constellation satellite navigation systems (GPS, Beidou, GLONASS, Galileo and QZSS). GNSS positioning can gain decent performance if GNSS receiver receive enough direct signals transmitted from satellites, so called line-of-sight (LOS). However, the GNSS propagation may be reflected, diffracted or blocked by skyscrapers and moving objects in super-urbanized area, such as Hong Kong, which can cause signal transmission delay. Thus, it introduces pseudorange errors due to both multipath effect and none-light-of-sight (NLOS) reception, which can present a positioning error of more than 100 meters in deep urban canyons [5].

Various researches are conducted to mitigate positioning errors caused by multipath effect and NLOS reception by designing specific GNSS receiver corrector [6, 7]. However, the GNSS receiver correlators can only be used to detect multipath which contains both direct signals and indirect signals. NLOS effects cannot be mitigated by correlators because NLOS only contains contaminated and reflected signals. Based on simulation of the possible GNSS signal transmission routes using the well-understood ray-tracing methods [8], 3D city maps aided (3DMA) GNSS [9-13] is developed to mitigate the multipath and exclude the NLOS receptions. Consistency check [14] methods are studied to detect the multipath effects without 3D city maps. Consistency of measurements between satellites is checked based on pseudorange residual. However, this technique may not provide satisfactory performance when there are numerous fake consistencies [15]. Vector tracking [16] method is also studied to mitigate multipath effects and detect the NLOS to improve the GNSS positioning. Effectiveness of the vector tracking-based multipath mitigation is also evaluated [17]. 3D laser scan is also used to construct the point cloud-based 3D geographic information of buildings, so-called the 3D point map. The 3D point cloud map is employed to detect the visibility of satellites [18]. To better model the reliability of GNSS positioning, horizontal dilution of precision (HDOP) is calculated using the reminding satellites and SNR is reconsidered to estimate the final covariance.

However, these previous studies, 3DMA GNSS, can only mitigate multipath effects and NLOS receptions caused by static buildings modelled in the 3D city maps. Moving objects with tall height, such as the double-decker bus [19] whose height can reach to 4.4 meters, can also cause NLOS receptions. In particularly, super-urbanized cities such as London and Hong Kong possessing numerous double-decker bus on the streets, which can introduce considerable errors into the pseudorange measurements. This GNSS positioning error caused by moving objects cannot be eliminated by the novel 3DMA GNSS. To obtain better GNSS positioning performance, this is an important issue that needed to be considered.

In this paper, we propose to exclude the NLOS receptions caused by moving objects in heavy traffic urban scenarios using real-time 3D point cloud generated by LiDAR. The multiple-channel LiDAR is widely used in autonomous driving vehicles [20, 21] and is employed to provide distance information of surrounding environments. Dimension and position of the dynamic object relative to GNSS receiver is calculated by object detection using the object detection and classification algorithms. Based on the detected objects boundaries, NLOS exclusion can be implemented with our proposed algorithm. Finally, GNSS positioning result is calculated based on the remaining visible satellites.

The remainder of this paper is structured as follows. An overview of the proposed method is given in Section II. Section III discusses double-decker bus detection method based on Euclidean clustering algorithm. Coordinate transformation of LiDAR coordinate system to skyplot coordinate system is also presented in this section. In Section IV, NLOS exclusion criterion is proposed and satellites exclusion is implemented. Then, GNSS WLS positioning is introduced. In Section V, we evaluate the effectiveness of the proposed method by means of experiments. Finally, a conclusions are withdrawn in Section VI.

## II. OVERVIEW OF THE PROPOSED METHOD

In this study, we focus on NLOS reception caused by double-decker bus, a representative moving object in Hong Kong. Fig. 1 presents direct propagation routes and potential NLOS reception of GNSS signal. The double-decker bus (height is 4.4 meters) can block signal transmitted from the satellite. Meanwhile, this GNSS signal is reflected by nearby building and finally received by GNSS receiver equipped on top of the autonomous vehicle, which results in NLOS reception.

As an essential sensor for positioning and perception of autonomous driving, 3D LiDAR (Velodyne 32) is installed on the top as shown in Fig. 1. In this study, LiDAR is employed to detect the surrounding double-decker buses. Then, NLOS exclusion is implemented based on detected double-decker boundaries parameters which are projected into a skyplot [22] with satellites. Finally, GNSS WLS positioning is conducted using the remaining satellites. The proposed method can be executed as follows:

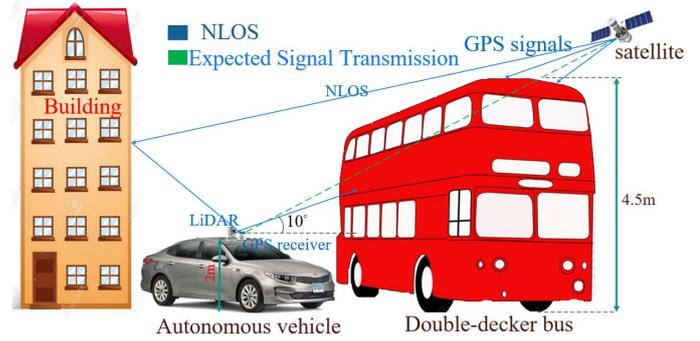

Fig. 1. Illustration of NLOS reception caused by double-decker bus.

Step I: Euclidean clustering is employed to transfer real-time 3D point clouds into several clusters. Parameters based classification method is utilized to classify the clusters and identify double-decker bus.

Step II: Satellites and double-decker bus are projected into a skyplot based on their azimuth and elevation relative to the GNSS receiver.

Step III: Considering satellites elevation, azimuth, SNR and double-decker bus boundary information (elevation and azimuth in skyplot), satellites which blocked by double-decker bus are excluded.

Step IV: Implementing GNSS WLS positioning using the surviving satellites after the Step III NLOS exclusion.

The details of the algorithms are introduced in the following sections.

## III. DOUBLE-DECKER BUS DETECTION AND TRANSFORMATION

Due to the limited field of view (+10° to -30° Vertical FOV), the LiDAR can only scan part of the double-decker bus body as shown in Fig. 1. In this section, Euclidean clustering [23] and parameters-based classification methods [24] are employed to detect the double-decker bus.

### A. Double-decker bus detection

From the view of LiDAR, the surrounding environment is represented as numerous points at a given time $t$ and the points are considered as a point set $P_t = \{p_1, p_2, ..., p_n, t\}$, where $p_i$ represents single point at a given time $t$. To give the points set $P_t$ a physical meaning, Euclidean clustering is implemented to divide it into several organized sets. The process of Euclidean clustering algorithm is summarized in detail as shown in algorithm 1.

Output of algorithm 1 is organized points sets $C_t^{clt} = \{C_1, C_2, ..., C_i, ... C_n, t\}$. To better portray the clusters, each cluster is represented by a descriptor, the bounding box $B$ [25]. Based on principle of bounding box, each $C_i$ in $C_t^{clt}$ can be transformed to $B_i$ in $B_t^{clt} = \{B_1, B_2, ..., B_i, ... B_n, t\}$ and is specifically determined by vector $B_i$ as follows:

$$B_i = [x_i^c, y_i^c, z_i^c, roll_i^c, pitch_i^c, yaw_i^c, d_i^{len}, d_i^{wid}, d_i^{al}] \quad (1)$$

where $x_i^c$, $y_i^c$ and $z_i^c$ denote the position of the bonding box in $x$, $y$, and $z$ directions respectively. $roll_i^c$, $pitch_i^c$ and $yaw_i^c$ denote the orientation of bounding box. $d_i^{len}$ is the length, $d_i^{wid}$ is the width and $d_i^{al}$ is the altitude of the bounding box. The bounding box list $B_t^{clt}$ contains both double-decker bus and other objects.

---

Algorithm 1: Euclidean clustering for points set $P_t$

---

**Input**: points set $P_t = \{p_1, p_2, ..., p_n, t\}$, search radius $r_{search}$
**Output**: organized points sets $C_t^{clt} = \{C_1, C_2, ..., C_i, ... C_n, t\}$
1  create a Kd-tree representation for the input points set $P_t$
2  set up an empty clusters list $C_t^{clt}$ and an empty list to save points sets $P_t^{check}$
3  **for all** points $p_i$ in $P_t$ **do**
4    add $p_i$ to the points set $P_t^{check}$
5    **for all** $p_i$ in $P_t^{check}$ **do**
6      search for the points set $C_i$ of point neighbor of $p_i$ in a sphere with radius r<$r_{search}$
7      **for** every point $C_i^i$ in points set $C_i$ **do**
8        **if** $C_i^i$ have not been processed
9          add $C_i^i$ to points sets $P_t^{check}$
10       **end if**
11     **end for** the points set $C_i$
12     **if** all the points in $P_t^{check}$ have been processed
13       add $P_t^{check}$ to $C_t^{clt}$
14       reset $P_t^{check}$ to empty
15     **end if**
16   **end for** $P_t^{check}$
17 **end for** $P_t$

---

To determine the double-decker bus clusters in bounding box list $B_t^{clt}$, parameters-based classification method is presented by the following three criterions.

$$\text{Classification}(B_i) = criter1 \&\& criter2 \&\& criter3 \quad (2)$$

The proposed three criterions are:

$$\begin{cases} criter1 = (d_i^{len} \in (len_{min}, len_{max})) \\ criter2 = (d_i^{wid} \in (wid_{min}, wid_{max})) \\ criter3 = (d_i^{al} \in (al_{min}, al_{max})) \end{cases} \quad (3)$$

where $len_{min}, len_{max}, wid_{min}, wid_{max}, al_{min}$ and $al_{max}$ are experimentally determined. If the value of Classification($B_i$) is 1, the bounding box is determined as a double-decker bus descriptor. As illustrated previously, only part of double-decker bus can be scanned by LiDAR which is represented by rectangle ABCD in Fig. 2. Dimension parameters of the bounding box representing double-decker bus can be extended to the real one in Hong Kong, whose length, width and height are 12.8, 2.5 and 4.4 meters respectively, which is represented by rectangle AEFB in Fig. 2. Then, the boundary parameter for the double-decker bus as shown in Fig. 2 is denoted by line segment $\overline{EF}$ denoted as $B_{bus}^{3d}$, the matrix of bus boundary. To represent the bus, two points, E and F, are required. The $B_{bus}^{3d}$ is structured as follows:

$$B_{bus}^{3d} = \begin{bmatrix} x_{3dE} & y_{3dE} & z_{3dE} \\ x_{3dF} & y_{3dF} & z_{3dF} \end{bmatrix} \quad (4)$$

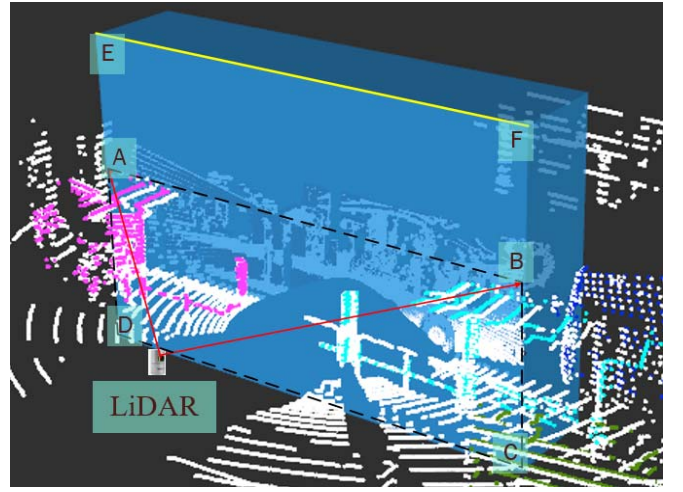

Fig. 2.  Illustration of Double-decker bus detection using Euclidean cluster algorithm and parameters-based classification. Blue box ABCD represents the initially detected double-decker bus. Blue box ABFE represents the extended detected double-decker bus.

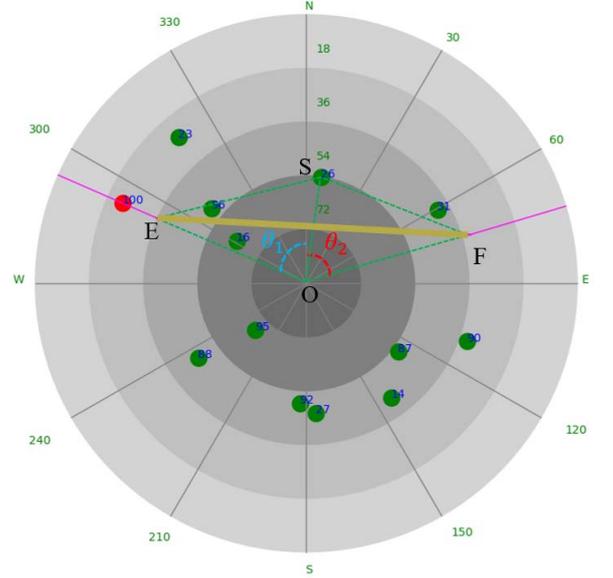

Fig. 3. Skyplot visualization for satellites and double-decker bus boundary. Green circles and the nearby numbers indicates satellites and corresponding PRNs. Line segment $\overline{EF}$ indicates the boundary.

Thus, the double-decker bus boundary is detected and transformed into the same coordinate system with the multi-constellation satellites.

*B. Coordination Transformation*

To implement the algorithm of NLOS exclusion, satellites' visibility needs to be determined based on the boundary of double-decker bus. Thus, satellites and double-decker bus parameters need to be transformed into a same coordinate, the skyplot. In each epoch, satellites information, including azimuth, elevation and SNR, can be acquired from the GNSS receiver. Satellites information can be represented as $SV_t^{all} = \{SV_1, SV_2, ..., SV_i, ... SV_n\}$. $SV_i$ represents the information for satellite $i$ and $SV_i = \{azim_i, ele_i, SNR_i\}$. $azim_i$ denotes the satellite azimuth. $ele_i$ represents satellite elevation and $SNR_i$ indicates satellite SNR.

Satellites can be easily transformed into skyplot (2-dimension coordinate) based on elevation and azimuth. Proper transformation matrix should be employed for double-decker bus boundary transformation from 3 dimensions coordinate to 2 dimensions coordinate. The transformation is conducted as the following formula.

$$B_{bus}^{skyp} = B_{bus}^{3d} G_T \quad (5)$$

where $B_{dou}^{3d}$ denotes the matrix of bus boundary presented in sub-section A. $G_T$ is a 3x2 transformation matrix. The $B_{dou}^{skyp}$ denotes the boundary matrix (2x2) in skyplot structured as follows:

$$B_{bus}^{skyp} = \begin{bmatrix} x_{skyE} & y_{skyE} \\ x_{skyF} & y_{skyF} \end{bmatrix} \quad (6)$$

After the transformation, satellites and double-decker bus can be presented in the same coordinate, the skyplot in Fig. 3. Line segment $\overline{EF}$ represents the double-decker bus boundary corresponding to line segment $\overline{EF}$ as shown in Fig. 2. Then, the azimuths for point E, and F can be calculated as $360 - \theta_1$ and $\theta_2$ respectively.

IV. IMPROVED GNSS POSITIONING BY NLOS EXCLUSION

In this section, NLOS exclusion criterion is proposed based on the detected double-decker bus boundary, satellites elevation, azimuth and signal to ratio (SNR). Then, GNSS positioning is conducted by WLS method.

A. NLOS Exclusion Based on Double-decker Bus Boundary

To exclude the satellites blocked by double-decker bus, relative position between each satellite and the detected bus boundary need to be calculated. As shown in Fig. 3, line segment $\overline{EF}$ represents the boundary of a double-decker bus. Satellite 26 (PRN 26) is located at point S. The azimuth and elevation are 8° and 54° respectively. The satellite exclusion procedure is summarized in detail as shown in algorithm 2. Inputs of algorithm 2 are the satellite information $SV_t^{all}$, bus boundary matrix $B_{bus}^{skyp}$, threshold of triangle area $S_{threshold}$, threshold of SNR $SNR_{threshold}$ and threshold of boundary uncertainty $\theta_{thres}$. Outputs of algorithm 2 are the survived satellites after NLOS exclusion. Firstly, angle $\theta_1$ and $\theta_2$ shown in Fig. 3 are estimated. Then areas of triangle $S_{\Delta SEO}$, $S_{\Delta SFO}$, $S_{\Delta SEF}$ and $S_{\Delta EOF}$ are calculated and $\Delta S$ can be estimated subsequently. Secondly, GNSS measurement that SNR is larger than $SNR_{threshold}$ will not be excluded. To avoid the faulty exclusion, a heuristically determined threshold $S_{threshold}$ is set. Satellites whose positions are quite near the extended edge beam ($\theta_1 < \theta_{thres}$ or $\theta_2 < \theta_{thres}$) also should not be excluded, such as the satellite 100 in Fig. 3. Satellites whose positions are quite near the double-decker bus boundary should not be excluded which can be judged by $\Delta S$, such as the satellite 31 and satellite 96 in Fig. 3. Finally, all the satellites in $SV_t^{all}$ are indexed and the satellites should not be excluded will be added to $SV_t^{surv}$. According to the

---

Algorithm 2: NLOS exclusion process

**Input**: Satellites information set $SV_t^{all} = \{SV_1, SV_2, ..., SV_i, ...SV_n\}$, bus boundary matrix $B_{bus}^{skyp}$, area threshold $S_{threshold}$, SNR threshold $SNR_{threshold}$, $\theta_{thres}$
**Output**: surviving satellites set after NLOS exclusion: $SV_t^{surv} = \{SV_1, SV_2, ..., SV_i, ...SV_m\}$
1 **for all** satellites $SV_i$ in $SV_t^{all}$ **do**
2    estimate $\theta_1, \theta_2$
3    Get triangle area $S_{\Delta SEO}$ of triangle SEO from $B_{bus}^{skyp}$
4    Get triangle area $S_{\Delta SFO}$ of triangle SFO from $B_{bus}^{skyp}$
5    Get triangle area $S_{\Delta SEF}$ of triangle SEF from $B_{bus}^{skyp}$
6    Get triangle area $S_{\Delta EOF}$ of triangle EOF from $B_{bus}^{skyp}$
7    $\Delta S = S_{\Delta SEO} + S_{\Delta SFO} + S_{\Delta SEF} - S_{\Delta EOF}$
8    **if** $(SNR_i > SNR_{threshold})$ or $(\theta_1 < \theta_{thres})$ or $(\theta_2 < \theta_{thres})$
9      **break**
10   **if** $\Delta S > S_{threshold}$ **and** $((\theta_1 + \theta_2) < \angle EOF < 180°$
11      **break**
12   **else**
13      add $SV_i$ to satellites set $SV_t^{surv}$
14   **end if**
15 **end for** satellites set $SV_t^{all}$

---

proposed NLOS exclusion algorithm in algorithm 2, satellites 23, 26 and 93 are going to be excluded.

After the NLOS exclusion process, satellites blocked by double-decker bus are excluded and can be employed to obtain better positioning performance.

B. GNSS Positioning Based on Surviving Satellites

Measurements with low elevation and SNR are more likely to be a contaminated GNSS signals, such as the multipath or NLOS, due to the reflection, blockage and diffraction. Thus, proper thresholds need to be set to exclude the unhealthy measurements. For satellite $SV_i$, if $ele_i$ is less than $ele_{thres}$ or $SNR_i$ is less than $SNR_{thres}$, it should be excluded from GNSS WLS positioning.

The clock bias between GNSS receiver and satellites is usually represented by the pseudorange measurement. The equation linking the receiver position and satellite can be structured as the following formula using least square (LS) method:

$$\hat{x} = (G^T G)^{-1} G^T \rho \quad (7)$$

where $G$ represents the observation matrix and is structured by unit LOS vectors between GNSS receivers position and satellites position. $\hat{x}$ indicates the estimated receiver position and $\rho$ denotes the pseudorange measurements.

To better represent the reliability of each measurement based on the information measured by receiver, weightings of each satellite are needed. Function to calculate the weighting by integrating the measurement SNR and satellite elevation is expressed as [22]:

$$W^{(i)}(ele_i, SNR_i) = \begin{cases} \frac{1}{\sin^2 ele_i} \left( 10^{-\frac{(SNR_i - T)}{a}} \left( \left( \frac{A}{10^{-\frac{(F-T)}{a}}} - 1 \right) \frac{(SNR_i - T)}{F - T} + 1 \right) \right) & SNR_i < T \\ 1 & SNR_i \geq T \end{cases} \quad (8)$$

where $W^{(i)}(ele_i, SNR_i)$ denotes the weighting for satellite $SV_i$. The parameter $T$ indicates the threshold of SNR and is equal to $SNR_{threshold}$. Parameter $a$, $A$ and $F$ in (8) are experimentally determined. Then, the weighting matrix W is a diagonal matrix constituted by the weightings $W^{(k)}(ele_i, SNR_i)$. Finally, GNSS receiver position can be estimated using WLS method as:

$$\hat{x} = (G^T W G)^{-1} G^T W \rho \quad (9)$$

Note that both LS (7) and WLS (8) positioning methods are compared in the experiment section.

## V. EXPERIMENT EVALUATION

### A. Experiment Setup

A static experiment is conducted near a bus stop in Hong Kong with lots of double-decker buses around. The ublox M8T receiver is used to collect raw GPS and Beidou measurements. 3D LiDAR sensor, Velodyne 32, is employed to provide the real-time point cloud. Both ublox receiver and 3D LiDAR are installed in a fix position near a static double-decker bus during the experiment which can be seen in Fig. 4. The data were collected at approximately 6 minutes at a frequency of 1 Hz. To verify the effectiveness of the proposed method, four methods were compared:

(1). LS positioning (LS)

(2). LS positioning + $ele_{thres}$ + $SNR_{thres}$ (LS-ESF)

(3). WLS positioning + $ele_{thres}$ + $SNR_{thres}$ (WLS-ESF)

(4). WLS positioning + $ele_{thres}$ + $SNR_{thres}$ + NLOS exclusion (WLS-ESF-NE)

In this experiment section, parameters mentioned above can be referrer in TABLE I.

### B. Comparision of Different GNSS Positioning Methods

Due to the double-decker bus is near the LiDAR sensor, boundary matrix $B_{bus}^{skyp}$ is always available throughout the static test. The experiment results of GNSS positioning using four methods are shown in TABLE II.

The LS method can achieve only 70.59 meters of mean errors among the test. Approximately 88.29 % of the results have a positioning error more than 40 meters. With the aid of elevation and SNR filters, the positioning error of LS-ESF decreases to 51.91 meters and about 63.24 % of the results possess a large error (> 40 meters). Meanwhile, the percentage of positioning error less than 20 meters is improved from 5.81 % to 11.3 %. This indicates that the elevation filter and SNR filter can enhance the positioning by excluding the unhealthy measurements. The reason behind this improvement is the exclusion of measurements 3, 91 and 22, which can be seen in Fig. 4. Those satellites possess low elevation, about 19°, are suffered from the severe NLOS/multipath effects, thus introducing considerable positioning errors. Slight improvement is obtained using WLS-ESF comparing with that of the LS-ESF method. The positioning error is decreased to 47.16 meters. This enhanced results indicates that weighting shown in (8) can effectively represents the health level for each measurements, thus an improved positioning result is acquired.

With the proposed NLOS exclusion method, the positioning results are considerably improved. Firstly, the positioning error and standard deviation (Std) of WLS-ESF-NE is reduced to 22.76 and 18.59 meters, respectively, comparing to that of WLS-ESF method. Secondly, almost 38 % of the results have a small positioning error (<20 meters). Moreover, Only 8.83 % of the results possess an error more than 40 meters. The reason for this improvement is the proposed NLOS exclusion as shown in Fig. 5. Satellites 23, 26 and 93 are excluded using the proposed algorithm 2. Though, the three satellites are blocked by double-decker bus, GNSS signals from them are reflected by surrounding buildings in the double-decker bus station, thus causing the erroneous NLOS receptions. The HDOP, positioning error and the numbers of measurement used in the WLS-ESF-NE and WLS-ESF method are shown in Fig. 6. The total satellites are over 10 all through the test, thus availability of GNSS positioning solution is 100 %. After the NLOS exclusion, HDOP value shown in the second panel is slightly increased, due to the change in the geometry distribution of satellites.

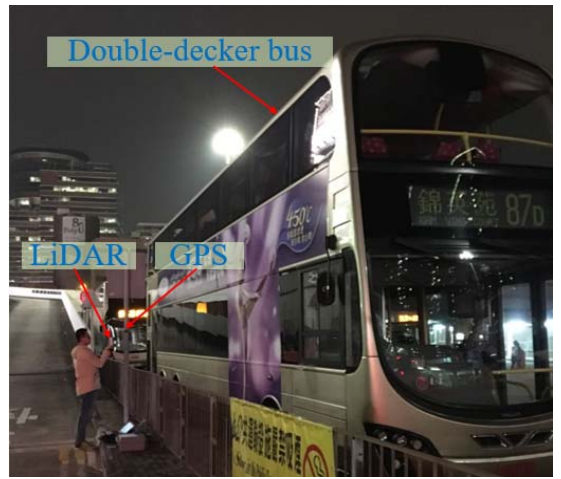

Fig. 4. Environemnt that the data were collected in a bus stop. Satellites can be blocked by the double-dekcer.

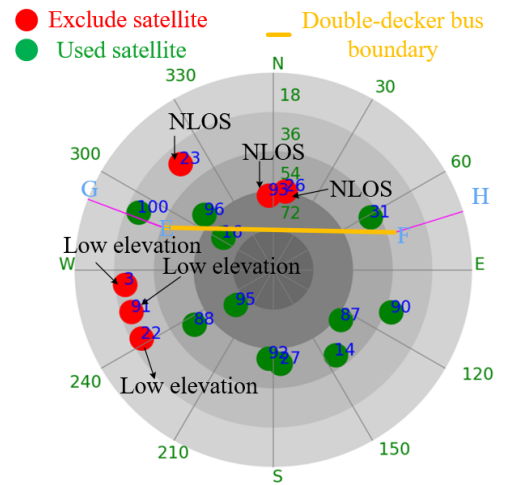

Fig. 5. Skyplot indicating the satellites distribution during the static experiment. Green circle represents the satellites that are healthy, which will be used in GNSS positioning. Red circle denotes the excluded satellites. Yellow line indicates the double-decker bus boundary.

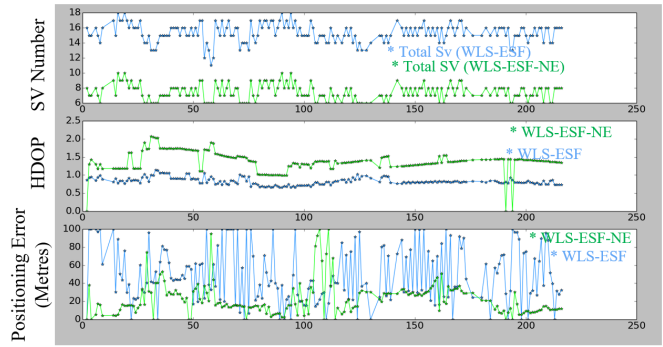

Fig. 6. Experimental results of WLS-ESF and WLS-ESF-NE, which depicted in blue and green dots, respectively. Top panel inciates the numbers of measurment used. Middle panles indicates the HDOP values. Button panels indicates the 3D positioning errors.

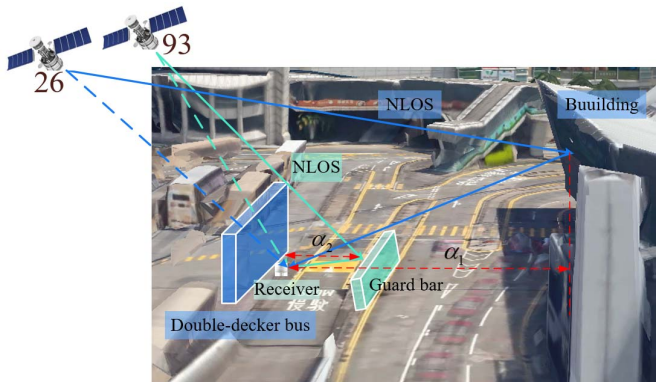

Fig. 7. GNSS signals transmission routes which causse NLOS receptions. Signal from satellite 26 is refelcted by building far away from receiver. Signal from satellite 93 is reflected by guard bar on the road side near receiver.

TABLE I.    PARAMETER VALUES USED IN THIS PAPER

| Parameters | $S_{threshold}$ | $SNR_{threshold}$ | $ele_{thres}$ | $\theta_{thres}$ |
|---|---|---|---|---|
| Value | 10 | 45 dB-Hz | 20° | 5° |
| Parameters | a | A | F | |
| Value | 30 | 32 | 10 | |

TABLE II.    POSITIONING PERFORMANCE OF THE FOUR METHODS NEAR A BUS STOP (IN THE UNIT OF METER)

| All data | LSP | LSP-ESF | WLSP-ESF | WLSP-ESF-NE |
|---|---|---|---|---|
| Mean error | 70.59 | 51.91 | 47.16 | 22.76 |
| Std | 26.0 | 29.4 | 32.34 | 18.59 |
| Percentage (<15 meters) | 5.81% | 11.35% | 14.58% | 38.00% |
| Percentage (<30 meters) | 9.12% | 28.11% | 34.46% | 77.61% |
| Percentage (>40 meters) | 88.29% | 63.24% | 50.14% | 8.83% |

TABLE III.    POSITIONING PERFORMANCE OF WLSP-ESF WITH SINGLE SATELLITE EXCLUSION (IN THE UNIT OF METER)

| All data | PRN23 | PRN26 | PRN93 | PRN100 |
|---|---|---|---|---|
| Mean error | 42.5m | 32.31m | 46.51m | 55.08m |
| Std | 27.53m | 26.67m | 30.01m | 30.28m |
| Percentage (<15 meters) | 15.33% | 18.51% | 6.38% | 4.12% |
| Percentage (<30 meters) | 37.71% | 67.72% | 38.29% | 25.13% |
| Percentage (>40 meters) | 45.90% | 23.28% | 45.74% | 59.16% |
| Improvement | 4.66m | 14.8m | 0.65m | 7.92m(worsen) |

*C. WLS-ESF Positioning with Manual Exclusion*

This sub-section presents the results of WLS-ESF with manual exclusion, meaning a specific measurement is excluded before using WLS-ESF method. TABLE II shows the results of four separated exclusion tests. Exclusion of satellite 23 introduces slight improvement in positioning performance with a mean error of 42.5 meters, comparing to the mean error of 47.16 meters using the WLS-ESF method without exclusion. As the GNSS signal received from satellite 23 is NLOS. Similarly, exclusion of satellites 26 and 93 also obtain improvements with a mean error of 32.31 meters and 46.51 meters respectively. The reason of this improvements distinction is that satellite 26 suffered larger NLOS errors comparing to satellites 93 which is subjected to the environments features. This can be seen in Fig. 7. According to [5], the NLOS delay in pseudorange domain is positive proportional to the ground distance from the receiver to the building that reflected the signal. Signals from satellites 26 and 93 are reflected by building and ground guard bar respectively. However, ground distance between receiver and the two separate reflectors are distinct ($\alpha_1$ for satellite 26, $\alpha_2$ for satellite 93). $\alpha_1$ is considerably larger than $\alpha_2$, therefore causing greater positioning error. On the contrary, greater improvement will be introduced if satellite 26 is excluded from GNSS positioning comparing with satellite 93.

After the exclusion of satellite 26, 67.72 % of the results possess an error less than 30 meters. However, exclusion of satellite 100 introduces larger positioning error comparing to the no exclusion situation. The mean error increases to 55.08 meters and approximately 59.16 % of the results possess an error more than 40 meters. The reason for this worsen performance is that satellite 100 actually is not blocked by double-decker bus though it is quite near the extended edge beam (line segment $\overline{EG}$ in Fig. 5). Thus, excluding satellites 23, 26 and 93 can all obtain improvements in GNSS positioning due to the double decker bus blockage and subsequent NLOS receptions.

*D. Discussion*

The evaluated four methods obtaining improved GNSS positioning performance based on more constraints are applied. Different satellites usually suffered from different range of positioning error (~30 meters) caused by NLOS receptions as shown in TABLE III. Satellite with high elevation can also be blocked by the double-decker bus, such as satellite 93 with an elevation of 54°. Meanwhile, low elevation does not equal to larger NLOS error, which can be referred by comparison between satellite 26 and 23, with an elevation of 53° and 27° respectively. Exclusion of satellite 26 obtained larger improvements with high elevation.

Moreover, improvement (from 47.16 to 22.76 meters, reduce by 24.4 meters) obtained by WLS-ESF-NE method is larger than the sum of improvement (PRN23:4.66 meters, PRN26:14.8 meters, PRN93:0.65 meters, total: 20.16 meters) introduced by WLS-ESF with manual exclusion. This is because the consistency of the pseudorange measurements improved after the exclusion of the unhealthy ones.

Similar to satellite 100, satellite 96, 16 and 31 also should not be excluded from GNSS positioning. Anyway, dimension extension of double-decker bus after detection is not absolute correct. As those three satellites are quite near the double-decker bus boundary with a lower possibility of being blocked. As shown in TABLE II, exclusion of satellite 100 can pose larger positioning error instead. Therefore, proper NLOS exclusion criterion is essential for obtaining better GNSS positioning.

As can be seen, the proposed method can exclude the satellites causing NLOS receptions and an improved GNSS positioning is obtained.

## VI. Conclusions and Future Work

With the rise of multi-constellation system, more satellites are available including GPS, BeiDou, GLONASS and Galileo. Number of visible satellite is still very enough for GNSS positioning even after NLOS exclusion. This study firstly employ object detection algorithm to detect a double-decker bus and extend its dimensions to a real one. Then, proper coordinate transformation is utilized to project double-decker bus boundary into GNSS skyplot. NLOS exclusion criterion using the elevation angle, SNR and bus boundary is proposed. According to the experiment result, the proposed method obtain best performance among the four conventional methods. The proposed method can effectively exclude the NLOS measurements and greatly enhance the positioning performance. With the aid of elevation and SNR filters, positioning performance is obvious improved which can be seen by comparing LS with LS-ESF method. The weighting scheme of measurement can slightly introduce improvement to the positioning performance. Positioning error of NLOS receptions caused by double-decker bus can reach 24 meters in overall. Finally, we conclude that exclusion of NLOS receptions is necessary for obtaining better GNSS positioning accuracy.

Furthermore, dynamic experiment will be conducted in urbanized area with complicated traffic conditions. The performance of the proposed method under dynamic scenarios will be further evaluated.

## Acknowledgment

The authors acknowledge the support of Hong Kong PolyU internal grant on the project G-YBWB, "Research on GPS Error Modelling Using 3D Point Cloud-Based Map for Autonomous Driving Vehicle".

## References

[1] C. Urmson et al., "Autonomous driving in urban environments: Boss and the Urban Challenge," (in English), *Journal of Field Robotics*, vol. 25, no. 8, pp. 425-466, Aug 2008.

[2] J. Levinson and S. Thrun, "Robust Vehicle Localization in Urban Environments Using Probabilistic Maps," (in English), *2010 Ieee International Conference on Robotics and Automation (Icra)*, pp. 4372-4378, 2010.

[3] Y. Gu, L.-T. Hsu, and S. Kamijo, "GNSS/onboard inertial sensor integration with the aid of 3-D building map for lane-level vehicle self-localization in urban canyon," *IEEE Transactions on Vehicular Technology*, vol. 65, no. 6, pp. 4274-4287, 2016.

[4] A. Fernandez et al., "GNSS/INS/LiDAR integration in urban environment: Algorithm description and results from ATENEA test campaign," in *Satellite Navigation Technologies and European Workshop on GNSS Signals and Signal Processing,(NAVITEC), 2012 6th ESA Workshop on*, 2012, pp. 1-8: IEEE.

[5] L.-T. Hsu, "Analysis and modeling GPS NLOS effect in highly urbanized area," *GPS solutions*, vol. 22, no. 1, p. 7, 2018.

[6] P. Fenton, "Pseudorandom noise ranging receiver which compensates for multipath distortion by making use of multiple correlator time delay spacing," ed: Google Patents, 1995.

[7] V. A. Veitsel, A. V. Zhdanov, and M. I. Zhodzishsky, "The mitigation of multipath errors by strobe correlators in GPS/GLONASS receivers," *GPS Solutions*, vol. 2, no. 2, pp. 38-45, 1998.

[8] Y. W. Lee, Y. C. Suh, and R. Shibasaki, "A simulation system for GNSS multipath mitigation using spatial statistical methods," (in English), *Computers & Geosciences*, vol. 34, no. 11, pp. 1597-1609, Nov 2008.

[9] P. D. Groves, "Shadow matching: A new GNSS positioning technique for urban canyons," *The journal of Navigation*, vol. 64, no. 3, pp. 417-430, 2011.

[10] M. Obst, S. Bauer, P. Reisdorf, and G. Wanielik, "Multipath Detection with 3D Digital Maps for Robust Multi-Constellation GNSS/INS Vehicle Localization in Urban Areas," (in English), *2012 Ieee Intelligent Vehicles Symposium (Iv)*, pp. 184-190, 2012.

[11] L. T. Hsu, Y. L. Gu, and S. Kamijo, "3D building model-based pedestrian positioning method using GPS/GLONASS/QZSS and its reliability calculation," (in English), *Gps Solutions*, vol. 20, no. 3, pp. 413-428, Jul 2016.

[12] T. Suzuki and N. Kubo, "Correcting GNSS Multipath Errors Using a 3D Surface Model and Particle Filter," (in English), *Proceedings of the 26th International Technical Meeting of the Satellite Division of the Institute of Navigation (Ion Gnss 2013)*, pp. 1583-1595, 2013.

[13] S. Miura, L. T. Hsu, F. Y. Chen, and S. Kamijo, "GPS Error Correction With Pseudorange Evaluation Using Three-Dimensional Maps," (in English), *Ieee Transactions on Intelligent Transportation Systems*, vol. 16, no. 6, Dec 2015.

[14] P. D. Groves and Z. Jiang, "Height Aiding, C/N-0 Weighting and Consistency Checking for GNSS NLOS and Multipath Mitigation in Urban Areas," (in English), *Journal of Navigation*, vol. 66, no. 5, pp. 653-669, Sep 2013.

[15] L.-T. Hsu, H. Tokura, N. Kubo, Y. Gu, and S. Kamijo, "Multiple Faulty GNSS Measurement Exclusion Based on Consistency Check in Urban Canyons," *IEEE Sensors Journal*, vol. 17, no. 6, pp. 1909-1917, 2017.

[16] L. T. Hsu, "Integration of Vector Tracking Loop and Multipath Mitigation Technique and Its Assessment," (in English), *Proceedings of the 26th International Technical Meeting of the Satellite Division of the Institute of Navigation (Ion Gnss 2013)*, pp. 3263-3278, 2013.

[17] L. T. Hsu, S. S. Jan, P. D. Groves, and N. Kubo, "Multipath mitigation and NLOS detection using vector tracking in urban environments," (in English), *Gps Solutions*, vol. 19, no. 2, pp. 249-262, Apr 2015.

[18] D. Maier and A. Kleiner, "Improved GPS Sensor Model for Mobile Robots in Urban Terrain," (in English), *2010 Ieee International Conference on Robotics and Automation (Icra)*, pp. 4385-4390, 2010.

[19] O. O. Okunribido, S. J. Shimbles, M. Magnusson, and M. Pope, "City bus driving and low back pain: A study of the exposures to posture demands, manual materials handling and whole-body vibration," (in English), *Applied Ergonomics*, vol. 38, no. 1, pp. 29-38, Jan 2007.

[20] J. Levinson et al., "Towards fully autonomous driving: Systems and algorithms," in *Intelligent Vehicles Symposium (IV), 2011 IEEE*, 2011, pp. 163-168: IEEE.

[21] J. Wei, J. M. Snider, J. Kim, J. M. Dolan, R. Rajkumar, and B. Litkouhi, "Towards a viable autonomous driving research platform," in *Intelligent Vehicles Symposium (IV), 2013 IEEE*, 2013, pp. 763-770: IEEE.

[22] J. Marshall, "Creating and viewing skyplots," *GPS solutions,* vol. 6, no. 1-2, pp. 118-120, 2002.

[23] R. B. Rusu, "Semantic 3d object maps for everyday manipulation in human living environments," *KI-Künstliche Intelligenz,* vol. 24, no. 4, pp. 345-348, 2010.

[24] P. Gehler and S. Nowozin, "On feature combination for multiclass object classification," in *Computer Vision, 2009 IEEE 12th International Conference on*, 2009, pp. 221-228: IEEE.

[25] G. Barequet and S. Har-Peled, "Efficiently approximating the minimum-volume bounding box of a point set in three dimensions," (in English), *Journal of Algorithms,* vol. 38, no. 1, pp. 91-109, Jan 2001.